\documentclass[10pt,twocolumn,letterpaper]{article}

\usepackage{wacv}
\usepackage{times}
\usepackage{epsfig}
\usepackage{graphicx}
\usepackage{amsmath}
\usepackage{amssymb}
\usepackage{dsfont}
\usepackage[english]{babel}

\DeclareMathOperator*{\argmax}{arg\,max}
\DeclareMathOperator*{\argmin}{arg\,min}

\def\wacvPaperID{****} % Enter the WACV Paper ID here

\wacvfinalcopy % *** Uncomment this line for the final submission

\ifwacvfinal
\def\assignedStartPage{1} % *** Enter the assigned starting page number (instead of 9876)
\fi

\ifwacvfinal
\usepackage[breaklinks=true,bookmarks=false]{hyperref}
\usepackage[capitalize]{cleveref}
\crefname{section}{Sec.}{Secs.}
\Crefname{section}{Section}{Sections}
\Crefname{table}{Table}{Tables}
\crefname{table}{Tab.}{Tabs.}

\addto\extrasenglish{%
}
\else
\usepackage[pagebackref=true,breaklinks=true,colorlinks,bookmarks=false]{hyperref}
\fi

\ifwacvfinal
\else
\fi

\begin{document}

%%%%%%%%% TITLE
\title{Less is More: Proxy Datasets in NAS approaches}

\author{Brian Moser$^{1,2}$, Federico Raue$^{1}$, Jörn Hees$^{1}$, Andreas Dengel$^{1,2}$\\
$^1$ German Research Center for Artificial Intelligence (DFKI), Germany\\
$^2$ TU Kaiserslautern, Germany\\
{\tt\small first.second@dfki.de}
}

\maketitle
%\thispagestyle{empty}

%%%%%%%%% ABSTRACT
\begin{abstract}
Neural Architecture Search (NAS) defines the design of Neural Networks as a search problem.
Unfortunately, NAS is computationally intensive because of various possibilities depending on the number of elements in the design and the possible connections between them.
In this work, we extensively analyze the role of the dataset size based on several sampling approaches for reducing the dataset size (unsupervised and supervised cases) as an agnostic approach to reduce search time.
We compared these techniques with four common NAS approaches in NAS-Bench-201 in roughly 1,400 experiments on CIFAR-100.
One of our surprising findings is that in most cases we can reduce the amount of training data to 25\%, consequently reducing search time to 25\%, while at the same time maintaining the same accuracy as if training on the full dataset.
Additionally, some designs derived from subsets out-perform designs derived from the full dataset by up to 22 p.p. accuracy.  
\end{abstract}

%%%%%%%%% BODY TEXT
\section{Introduction}
In recent years, a novel field called Neural Architecture Search (NAS) has gained interest, which aims to automatically find designs instead of hand-designed Neural Networks (NNs) created by researchers based on their knowledge and experience \cite{ren2020comprehensive}.
For example, the NAS approach AmoebaNet (developed by Google) reached state-of-the-art performances for the ImageNet Classification Task \cite{deng2009imagenet, real2019regularized}. 
Despite promising results, the main drawback is the computation time (especially for large datasets) that NAS approaches require to derive an architecture.
In addition, regular weight optimization of found architecture designs is still necessary to evaluate the quality of design choices.
For instance, this is the case for AmoebaNet, which selects different configurations via trial-and-error as an evolutionary approach. 
Thus, the selection requires training of each configuration to evaluate the fitness.
Therefore, researchers tend to constrain the search space of a given NAS algorithm as a trade-off to runtime speed \cite{liu2018darts, pham2018efficient, dong2019searching, suganuma2017genetic, ren2020comprehensive}.

Nevertheless, NAS approaches sometimes use datasets that are sub-optimal for NAS as a whole.
In more detail, the role of each sample is not always positive and can even hurt the performance, which is observable for datasets used for Image Classification tasks like ImageNet \cite{deng2009imagenet, shleifer2019using, katharopoulos2018not}.
With this in mind, we are interested in analyzing the role of the training dataset size as an approach to reducing the search time in NAS.
Thus, this work evaluates several sampling methods for selecting a subset of a dataset for supervised and unsupervised scenarios with four NAS approaches from NAS-Bench-201 \cite{dong2020bench}.
We evaluated on CIFAR-100 \cite{cifar100} that the NAS approach DARTS \cite{liu2018darts} derived an architecture with 53.75\% top-1 accuracy and a search time of 54 hours as a baseline on an RTX 2080 GPU by NVIDIA. In contrast, it was possible to reach 75.20\% top-1 accuracy within a search time of just 13 hours on 25\% of the training data with the same NAS approach. Furthermore, for another NAS approach, GDAS \cite{dong2019searching}, it was possible to derive an architecture with comparable results on a 50\% reduced subset compared to the baseline.
The contributions of this work are
\begin{itemize}
 \item Evaluation of six different sampling methods for NAS, divided into three supervised and three unsupervised methods. The evaluation was done with ca. 1,400 experiments on CIFAR-100 with four NAS algorithms from NAS-Bench-201 (DARTS-V1 \cite{liu2018darts}, DARTS-V2 \cite{liu2018darts}, ENAS \cite{pham2018efficient}, and GDAS \cite{dong2019searching}).
 \item Improvement of NAS search time by using 25~\% of the dataset, resulting in only 25~\% of computation time, parallel to better cell designs that outperformed the baseline, sometimes by a large margin (22 p.p.).
 \item Explanation of performances by detailed investigation of design choices taken by the NAS algorithms and showing the generalizability with ImageNet-16-120.
\end{itemize}

\section{Related Work}
For this work, two related areas are essential to be described.
The first area is the role of each sample of the dataset on model performance.
We use the idea of reducing the dataset size as an approach to scale down searching time in NAS.
The second area is benchmarking NAS approaches for Image Classification using a common framework called NAS-Bench-201 to evaluate different sampling methods.

%-------------------------------------------------------------------------
\subsection{Proxy Datasets}
A \textit{proxy} usually refers to an intermediary in Computer Science \cite{gamma1995elements}.
In our case, a proxy dataset $D_r$ is an intermediary for the original dataset $D$ and the NAS search phase.
Mathematically, a proxy dataset is a subset of the original dataset $\mathcal{D}_r \subset \mathcal{D}$ with a size-ratio $r \in (0,1)$.
So far, the proxy dataset concept was shown to be successful in Image Classification tasks \cite{shleifer2019using, katharopoulos2018not}.
There are two ways of creating such proxy datasets.
One way is to generate syntactic samples which represent a compressed version of the original dataset.
\textit{Dataset Distillation} \cite{wang2018dataset} and \textit{Dataset Condensation} \cite{zhao2021dataset} propose similar approaches in which NNs train on small-sized datasets of synthetic dataset (e.g., 10 or 30 samples per class) and reach better results than using real samples from the original datasets.
Another way is to select only training samples that are beneficial for training.
Schleifer et al. \cite{shleifer2019using} proposed a hyper-parameter search on proxy datasets such that experiments on the proxies highly correlate with experimental results on the entire dataset.
Therefore, the hyper-parameter search can be performed faster on a proxy dataset without forfeiting significant performance on the complete dataset. 

In this work, we explore the second approach for selecting samples from training data during the search for architecture designs.
Our goal is to compare several sampling methods that derive proxy datasets and speed up NAS approaches by reducing the training dataset size (i.e., computation-intensive Cell-Based Search).

\subsection{NAS-Bench-201}
\label{chapter_nas_bench}
One major problem in NAS research is how hard it is to compare NAS approaches due to different spaces, e.g., unalike macro skeletons or sets of possible operations for an architecture \cite{zoph2018learning, tan2019mnasnet, pham2018efficient}.
Additionally, researchers use different training procedures, such as hyper-parameters, data augmentation, and regularization, which makes a fair comparison even harder \cite{liu2018progressive, ying2019bench, dong2019one, dong2020bench}.
Therefore, Dong et al. \cite{dong2020bench} released \textit{NAS-Bench-201} (an extension of NAS-Bench-101 \cite{ying2019bench}) that supports fair comparisons between NAS approaches.
The benefit of using NAS-Bench-201 is efficiently concluding the contributions of various NAS algorithms.
Its set of operations $\mathcal{O}$ provides five types of operations that are commonly used in the NAS literature: $\mathcal{O} = \{ {3\times3\text{ Conv}}, {3\times3\text{ Avg}}, {1\times1\text{ Conv}},{\text{ Skip}},{\text{Zeroize}} \}$.

In this work, we exploit NAS-Bench-201 for comparing several NAS approaches applied to proxy datasets.

%-------------------------------------------------------------------------

\section{Methodology}
\label{chapter_sampling}

Our work relies on several sampling methods for extracting proxy datasets.
Additionally, we consider sampling approaches for both scenarios, supervised and unsupervised, as will be described in the following sections.
% We describe the following approaches for \textbf{unsupervised} sampling: Random Sampling (RS), K-Means Outlier Removal (KM-OR), and sampling via an Autoencoder (AE).
% We describe the following approaches for \textbf{supervised} sampling: Class-Conditional Random Sampling (CC-RS), Class-Conditional Outlier Removal (CC-OR), and sampling via Transfer Learning (TL).

\subsection{Proxy Datasets and Sampling}
Let $\mathcal{D} = \{ \left( \mathcal{X}_i, y_i \right) \}$ be a dataset with cardinality $n_\mathcal{D}$. 
%Throughout this work, the training set is used during the Cell Search. 
$\mathcal{X}_i$ with $0 \leq i < n_\mathcal{D}$ denotes the $i$-th sample of the dataset and $y_i$ its corresponding label, if it exists (supervised case).
For the dataset, $\mathcal{C} = \{ \mathcal{C}_j \}$ with cardinality $n_\mathcal{C}$ denotes the labels. 
A subset $\mathcal{D}_r \subset \mathcal{D}$ will be called proxy with cardinality $n_{\mathcal{D}_r}$. 
The ratio $r \in \left( 0, 1 \right)$ denoted as index here indicates the  remaining percentage size of the original dataset $\mathcal{D}$: $n_{\mathcal{D}_r} \approx r \cdot n_\mathcal{D}$ (approximate because the datasize is not always divisible without remainder). 
The proxy dataset operates between the original dataset and the NN.
In this work, defining a proxy dataset aims to decrease the time needed for experiments without suffering a quality loss compared to a run on the entire dataset.
Ideally, the proxy dataset should also improve the quality of the NAS design choices.

The goal is to derive a proxy dataset $\mathcal{D}_r \subset \mathcal{D}$ with cardinality $n_{\mathcal{D}_r} \approx r \cdot n_\mathcal{D}$, $r \in \left( 0, 1 \right)$.
Thus, for any given $r \in \left( 0, 1\right)$, the sampling method has to ensure 

\begin{equation}
    n_{\mathcal{D}_r} = \sum_{ \left(\mathcal{X}_i, y_i \right) \in \mathcal{D}} \mathds{1}_{\mathcal{D}_r} \left( \mathcal{X}_i \right) \approx r \cdot n_\mathcal{D},
    \label{eq:sampling}
\end{equation}

\noindent
where $\mathds{1}_{\mathcal{D}_r}: \mathcal{D} \rightarrow \{ 0, 1\}$ is the indicator function that indicates the membership of an element in a subset $\mathcal{D}_r$ of $\mathcal{D}$.

\subsection{Unsupervised Sampling}
Unsupervised sampling methods do not take the label $y_i$ of a training sample into account for sampling.

\subsubsection{Random Sampling (RS)}
Each sample of the dataset has an equal probability of being chosen.
Therefore, it holds for $i \neq j$ that $\mathbb{P} \left[ \mathds{1}_{\mathcal{D}_r} \left( \mathcal{X}_i \right) = 1\right] = \mathbb{P} \left[ \mathds{1}_{\mathcal{D}_r} \left( \mathcal{X}_j \right) = 1\right]$.
In consequence, for any $r \in \left( 0, 1\right)$, one can derive a randomly composed subset $D_r \subset D$ such that 
\begin{equation}
    \sum_{\left(\mathcal{X}_i, y_i \right) \in \mathcal{D}} \mathbb{P} \left[ \mathds{1}_{\mathcal{D}_r} \left( \mathcal{X}_i \right) = 1\right] \approx r \cdot n_\mathcal{D}
    \label{eq:rs_1}
\end{equation}

\subsubsection{K-Means Outlier Removal (KM-OR)}
An \textit{outlier} is a data point that differs significantly from the leading group of data. While there is no generally accepted mathematical definition of what constitutes an outlier, it is straightforward to define outliers for any $r \in \left( 0, 1 \right)$ as the $(1-r) \cdot n_{\mathcal{D}}$ sample points that have the highest cumulative distance to its group centers in the context of this work.
In order to identify groups, one can use the K-Means clustering algorithm.
Thus, it is possible to derive a proxy dataset by removing outliers from each cluster. Let $d$ be a distance metric, e.g., the Frobenius norm.
Given the cluster centroids $S = \{\mu_1, ..., \mu_K\}$ of K-Means and $r \in \left( 0, 1 \right)$, the derived proxy dataset is then
\begin{equation}
    \mathcal{D}_r = \argmin_{\substack{\mathcal{D}' \subset \mathcal{D}, \\ \text{s.t. } \mid \mathcal{D}'\mid \approx r \cdot n_{\mathcal{D}}}} \ \sum_{\left(\mathcal{X}_i, y_i\right) \in \mathcal{D}'} \min_{\mu_j \in S} d \left( \mu_j, \mathcal{X}_i \right),
    \label{eq:kmeans_outlier}
\end{equation}
\subsubsection{Loss-value-based sampling via AE (AE)}
The typical use case of an Autoencoder (AE) is to approximate $X_i \approx D \left( E \left( X_i \right) \right)$, where $E$ is an encoder and $D$ a decoder of an AE \cite{kramer1991nonlinear}.
Given a trained AE, it can provide a distance metric based on a loss function.
Let $\mathcal{L}: \mathbb{R}^{h \times w \times c} \times \mathbb{R}^{h\times w \times c} \rightarrow \mathbb{R}$ be a loss function like Mean Squared Error, with $h$, $w$, $c$ as the height, width, and channel size, respectively.
Given $r \in \left( 0, 1\right)$, one can derive a proxy dataset with
\begin{equation}
    \mathcal{D}_r = \argmin_{\substack{\mathcal{D}' \subset \mathcal{D}, \\ \text{s.t. }\mid \mathcal{D}'\mid \approx r \cdot n_\mathcal{D}}} \sum_{\left(\mathcal{X}_i, \cdot \right) \in \mathcal{D}'} \mathcal{L} \left( D \left( E \left( \mathcal{X}_i \right)\right), \mathcal{X}_i\right),
\end{equation}
\noindent
where samples with high loss-values are removed, which are typically the hardest samples to reconstruct for the AE.
The opposite direction ($\argmax$ instead of $\argmin$) was tested with less significant results.
It can be found in the supplemental material \cite{suppCVPR22}.

\subsection{Supervised Sampling}
Supervised sampling methods take the label $y_i$ of a training sample into account for sampling.

\subsubsection{Class-Conditional Random Sampling (CC-RS)}
Each sample of a class of the dataset has an equal probability of being chosen. 
In contrast to pure Random Sampling, the class is considered to ensure an equal sampling within each class:
\begin{equation}
\begin{split}
        & \sum_{\left(\mathcal{X}_i, y_i \right) \in \mathcal{D}} \mathbb{P} \left[ \mathds{1}_{\mathcal{D}_r} \left( \mathcal{X}_i \right) = 1\right] \approx r \cdot n_\mathcal{D} \\
        \text{s.t. } & P \left[ \mathds{1}_{\mathcal{D}_r} \left( \mathcal{X}_i \right) = 1 \mid y_i = \mathcal{C}_j \right] = \mathbb{P} \left[ \mathds{1}_{\mathcal{D}_r} \left( \mathcal{X}_i \right) = 1\right] \\
        &\forall \mathcal{C}_j \in \mathcal{C}
\end{split}
\end{equation}

\subsubsection{Class-Conditional Outlier Removal (CC-OR)}
\label{cc_outlier_removal_sec}
Similar to K-Means Outlier Removal, one can use the class centroids in order to define clusters.
%In an ideal case, each class contains samples that form a cluster.
Therefore, it is possible to derive a proxy dataset by removing outliers from each class.
Let $\mathcal{C}_j \in \mathcal{C}$ be a class, $\Omega_{\mathcal{C}_j}$ be the data points that lie within the class $\mathcal{C}_j$ and $\mu_{\mathcal{C}_j}$ the class centroid of $\mathcal{C}_j$. Collecting the samples to derive a proxy dataset $\mathcal{D}_r$ with $r \in \left( 0, 1 \right)$ can be defined as 
\begin{equation}
    \mathcal{D}_r = \bigcup_{\mathcal{C}_j \in \mathcal{C}} \ \argmin_{\substack{\mathcal{D}' \subset \Omega_{\mathcal{C}_j}, \\ \text{s.t. }\mid \mathcal{D}'\mid \approx r \cdot n_{\mathcal{C}_j}}} \ \sum_{\left(\mathcal{X}_i, y_i\right) \in \mathcal{D}'} d \left( \mu_{\mathcal{C}_j}, \mathcal{X}_i \right),
    \label{eq:cc_outlier}
\end{equation}

\subsubsection{Loss-value-based sampling via Transfer Learning (TL)}
\label{tl_sampling}
Instead of using centroids or AE, one can use Transfer Learning (TL) to derive a classifier $\varphi$ that enables a loss-value-based distance metric.
Hence, it gives the possibility to obtain a proxy dataset of $\mathcal{D}$ with the easiest to classify samples. 
Given $r \in \left( 0, 1\right)$, one can derive such a proxy dataset, where samples with high loss-values are removed:
\begin{equation}
    \mathcal{D}_r = \argmin_{\substack{\mathcal{D}' \subset \mathcal{D}, \\ \text{s.t. }\mid \mathcal{D}'\mid \approx r \cdot n_\mathcal{D}}} \sum_{\left(\mathcal{X}_i, y_i\right) \in \mathcal{D}'} \mathcal{L} \left( \varphi \left( \mathcal{X}_i \right), y_i\right),
\end{equation}
\noindent
In the following, we use a classifier $\varphi$ trained on ImageNet and Transfer Learning it to CIFAR-100.
Similar to the unsupervised case AE, the opposite direction ($\argmax$ instead of $\argmin$) was tested with less significant results. 
It can be found in the supplemental material \cite{suppCVPR22}.

\section{Experiments}
This Section introduces the CIFAR-100 dataset and the evaluation strategy used in this work.
Then, it continues with the quantitative results, which show the performance of all experiments and the time savings, and ends with the qualitative results.
The code for our experiments can be found on GitHub\footnote{\url{https://github.com/anonxyz123/fgk62af5/}}.
For more details, the supplemantal material \cite{suppCVPR22} lists the hyper-parameters for the experiments as well as other experimental results, e.g., alternative loss-value-based sampling methods or additional $K$-values for K-Means Outlier Removal.

\subsection{Data Set}
The dataset $\mathcal{D}$ used in this work is CIFAR-100 \cite{cifar100}, which is a standard dataset used for benchmarking NAS approaches because of its size and complexity compared to CIFAR-10 \cite{cifar10} and SVHN \cite{Netzer2011}.
It contains $n_\mathcal{C}=100$ classes and has a uniform distribution of samples between all classes ($n^{train}_{\mathcal{C}_j}=500$ training and $n^{test}_{\mathcal{C}_j}=100$ testing samples $\forall \mathcal{C}_j \in \mathcal{C}$). 
%Each sample $\left(\mathcal{X}_i, y_i\right) \in \mathcal{D}$ consists of $\mathcal{X}_i \in \mathbb{R}^{32\times32\times3}$, a color image, and a label, where $\exists j \in \{ 0, n_\mathcal{C} - 1 \}: y_i = \mathcal{C}_j$. 
In total, $\mathcal{D}$ has $n_\mathcal{D} = 60,000$ samples.

\subsection{Evaluation Strategy}
\label{sec:eval}

\begin{figure}
\begin{center}
    \includegraphics[width=.8\linewidth]{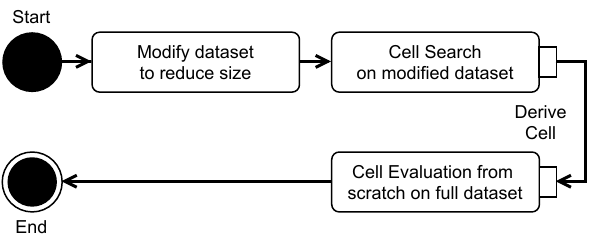}
\end{center}
\caption{Evaluation Process. First, sampling is applied to derive a proxy dataset. After that, this work applies a NAS algorithm to the reduced proxy dataset. Next, the derived Cell is trained from scratch on the full dataset.}
\label{fig:eval_proc}
\end{figure}

This work uses NAS-Bench-201\footnote{\url{https://github.com/D-X-Y/AutoDL-Projects/}} (MIT licence) as a framework for the search space as discussed in~\autoref{chapter_nas_bench}.
The NAS algorithms used in this work are DARTS (first-order approximation V1 and second-order approximation V2) \cite{liu2018darts}, ENAS \cite{pham2018efficient}, and GDAS \cite{dong2019searching}.
The sample ratios are defined by $r \in \{ 0.25, 0.50, 0.75, 1.0 \}$, where $r=1.0$ means that the whole dataset is used.
%Let $R = \{ 0.25, 0.50, 0.75, 1.0 \} \subset \left( 0, 1 \right)$ be the sample ratios with $r \in R$.
Given any sampling method, the proxy dataset $\mathcal{D}_r$ is selected once as discussed in~\autoref{chapter_sampling} and evaluated with the previously-mentioned NAS algorithms for a fair comparison. 
We evaluate the sampling methods on the proxy dataset $\mathcal{D}_r$ based on the NAS algorithm, which returns a cell design.

Two processes are essential in our experimental setup: \textit{Cell Search} and \textit{Cell Evaluation}.
The \textit{Cell Search} uses the proxy dataset and is applied once for all NAS approaches and sampling methods.
Additionally, we have fixed the channel size of each convolution layer in the cell to 16 since it is suggested by the NAS-Bench-201 framework. 
Afterwards, the \textit{Cell Evaluation} process starts: A macro skeleton network uses the cell design and is trained from scratch on the original dataset $\mathcal{D}$.
This is repeated \textit{three times} with different weight initialization for evaluating the robustness of the cell design.
Thus, the results report a mean and standard deviation value. 
\autoref{fig:eval_proc} illustrates the proxy dataset sampling and the processes taken afterward. 
Compared to the default setup of NAS-Bench-201, where the Cell Evaluation is done with one fixed channel size, this work extended the evaluation for different channel sizes (16, 32, and 64) to survey the scalability of the found cell designs. 

%In conclusion, each sampling method has four Cell Searches because of four tested NAS approaches for three different $r$ settings, resulting in twelve experiments with twelve cell designs.
In summary, each sampling method is applied under two conditions.
The first condition is the size of the dataset after applying the sampling method that is three different $r$ settings.
The second condition is the NAS approaches, which are four in this work.
Furthermore, the Cell Evaluation applies three different channel sizes repeated three times with non-identical starting weights, which results in nine experiments for each design choice.
Thus, we are running $(4\cdot3) + (4\cdot3) \cdot (3 \cdot 3) = 120$ experiments to evaluate an individual sampling method.
This work presents the six sampling methods (listed in~\autoref{chapter_sampling}). 
Additionally, we evaluate other five sampling approaches (two alternative loss-value-based formulations and three other K-Values for K-Means) for completeness, and they are presented in the supplemental material \cite{suppCVPR22}.
As a result, we run roughly 1,400 experiments ($11 \cdot 120$ + baselines).

\subsection{Quantitative Analysis}
This Section describes and analyzes the results of all four NAS algorithms applied on sampled proxy datasets (as discussed in~\autoref{chapter_sampling}).

\begin{table}
\begin{center}
    \caption{Cell Evaluation with \textbf{DARTS-V1} (all accuracy values in percent). The poorly performing so called ``local optimum cell design'' consisting only of Skip-Connections (discussed in~\autoref{localOptimumCell}) is marked with *. It is found on randomly sampled proxy datasets. Besides that, cell designs derived in all other unsupervised cases reach similar results like the baseline. However, cell designs for the supervised case outperform the baseline by a large margin (7.3 p.p. up to 34 p.p. for C=16). Note also that manually adding more channels (C=32 and C=64) to the found cell design has a positive effect. The performance improvement gained by adding more channels seems to be consistent across all experiments.}
    
    \label{table_dartsv1}
    \begin{tabular}{l  c  c  c  c}
        \hline
      Method & $r$ & A, $C$=16 & A, $C$=32 & A, $C$=64 \\
      \hline
      \multicolumn{2}{l}{\textbf{Baseline}}& & & \\
      Full DS & 1.0 & 32.6 $\pm$ 0.7 & 40.2 $\pm$ 0.4 & 47.3 $\pm$ 0.3 \\
      \hline
      \multicolumn{2}{l}{\textbf{Unsupervised}}& & & \\
      \textbf{RS} & \textbf{.75} & \textbf{40.0 $\pm$ 0.1} & \textbf{50.9 $\pm$ 0.5} & \textbf{55.2 $\pm$ 0.5} \\
      RS* & .50 & 15.9 $\pm$ 0.7 & 17.7 $\pm$ 0.2 & 18.3 $\pm$ 0.1 \\
      RS* & .25 & 15.9 $\pm$ 0.7 & 17.7 $\pm$ 0.2 & 18.3 $\pm$ 0.1 \\
      KM-OR & .75 & 35.2 $\pm$ 0.2 & 44.8 $\pm$ 0.3 & 51.9 $\pm$ 0.7 \\
      KM-OR & .50 & 27.3 $\pm$ 0.8 & 33.1 $\pm$ 0.2 & 38.5 $\pm$ 0.4 \\
      \textbf{KM-OR} & \textbf{.25} & \textbf{40.0 $\pm$ 0.1} & \textbf{51.0 $\pm$ 0.5} & \textbf{55.3 $\pm$ 0.5} \\
      AE & .75 & 34.1 $\pm$ 0.3 & 42.6 $\pm$ 0.2 & 50.6 $\pm$ 0.0 \\
      \textbf{AE} & \textbf{.50} & \textbf{66.6 $\pm$ 0.5} & \textbf{72.4 $\pm$ 0.3} & \textbf{75.7 $\pm$ 0.1} \\
      AE & .25 & 33.3 $\pm$ 0.4 & 42.3 $\pm$ 0.2 & 50.6 $\pm$ 0.3 \\
      \hline
      \multicolumn{2}{l}{\textbf{Supervised}}& & & \\
      CC-RS* & .75 & 15.9 $\pm$ 0.7 & 17.7 $\pm$ 0.2 & 18.3 $\pm$ 0.1 \\
      CC-RS* & .50 & 15.9 $\pm$ 0.7 & 17.7 $\pm$ 0.2 & 18.3 $\pm$ 0.1 \\
      CC-RS* & .25 & 15.9 $\pm$ 0.7 & 17.7 $\pm$ 0.2 & 18.3 $\pm$ 0.1 \\
      CC-OR & .75 & 63.8 $\pm$ 0.4 & 70.4 $\pm$ 0.3 & 74.9 $\pm$ 0.5 \\
      \textbf{CC-OR} & \textbf{.50} & \textbf{64.1 $\pm$ 0.3} & \textbf{70.4 $\pm$ 0.2} & \textbf{75.4 $\pm$ 0.3} \\
      CC-OR & .25 & 54.6 $\pm$ 0.2 & 59.1 $\pm$ 0.4 & 60.8 $\pm$ 1.1 \\
      TL & .75 & 60.6 $\pm$ 0.4 & 62.3 $\pm$ 0.5 & 64.6 $\pm$ 0.3 \\
      \textbf{TL} & \textbf{.50} & \textbf{60.6 $\pm$ 0.2} & \textbf{62.8 $\pm$ 0.4} & \textbf{65.0 $\pm$ 0.5} \\
      TL & .25 & 39.9 $\pm$ 0.4 & 51.3 $\pm$ 0.2 & 57.0 $\pm$ 0.5 \\
      \hline
    \end{tabular}
\end{center}
\end{table}
\subsubsection{DARTS-V1}
\autoref{table_dartsv1} shows the Cell Evaluation for DARTS-V1. 
As mentioned in~\autoref{sec:eval}, the Cell Search was done with a channel size of 16, and the found cell design was evaluated by training it from scratch on the full dataset with channel sizes of 16, 32, and 64.
Besides RS and CC-RS, the proxy datasets worked well for DARTS-V1.
As can be observed, even the worst accuracy results among proxy datasets derived by loss-value-based (AE \& TL) sampling achieved similar results as the baseline with a possibility to reduce the size to 25\%.
Additionally, it was possible to outperform the baseline on 25\% of the dataset size with CC-OR. The best performance gain (AE, $r=0.5$) achieved a margin of +28.4 p.p. compared to baseline.
Also, one can observe that increasing the channel sizes consistently increases the network performance. 
Regarding RS, DARTS-V1 and the following NAS algorithms derive a cell design consisting only of Skip-Connections (marked with *). It explains the bad accuracy during Cell Evaluation because it has no learnable parameters. 
This happens due to instability within DARTS, which is known in literature \cite{bi2019stabilizing} and discussed in~\autoref{localOptimumCell}.

\subsubsection{DARTS-V2}
\autoref{table_dartsv2}  shows the Cell Evaluation for DARTS-V2, which is similar to DARTS-V1. However, the evaluation on all loss-value-based (AE \& TL) sampled proxy datasets delivers better results than the baseline. The best performance gain (TL, $r=0.5$) achieved a margin of +22 p.p. compared to baseline.
The AE approach even gets better with decreasing dataset size. 
The observation of good results also holds for CC-OR. 
For KM-OR and CC-RS, the results are close to the baseline if the cell design with only Skip-Connections is not derived.
Nevertheless, the proxy dataset derived by RS concludes the aforementioned bad-performing design for all $r$-values.

\begin{table}
\begin{center}
    \caption{Cell Evaluation with \textbf{DARTS-V2}. The local optimum cell design (discussed in~\autoref{localOptimumCell}) is marked with *. The results are similar to DARTS-V1, slightly better for AE, TL, and CC-OR.}
    \label{table_dartsv2}
    \begin{tabular}{l  c  c  c  c}

    \hline
      Method & $r$ & A, $C$=16 & A, $C$=32 & A, $C$=64 \\
      \hline
      \multicolumn{2}{l}{\textbf{Baseline}}& & & \\
      Full DS & 1.0 & 35.7 $\pm$ 0.5 & 46.0 $\pm$ 0.3 & 53.7 $\pm$ 0.4 \\
      \hline
      \multicolumn{2}{l}{\textbf{Unsupervised}}& & & \\
      RS* & .75 & 15.9 $\pm$ 0.7 & 17.7 $\pm$ 0.2 & 18.3 $\pm$ 0.1 \\
      RS* & .50 & 15.9 $\pm$ 0.7 & 17.7 $\pm$ 0.2 & 18.3 $\pm$ 0.1 \\
      RS* & .25 & 15.9 $\pm$ 0.7 & 17.7 $\pm$ 0.2 & 18.3 $\pm$ 0.1 \\
      KM-OR & .75 & 32.1 $\pm$ 0.6 & 39.7 $\pm$ 0.4 & 46.4 $\pm$ 0.2 \\
      KM-OR* & .50 & 15.9 $\pm$ 0.6 & 17.7 $\pm$ 0.2 & 18.3 $\pm$ 0.1 \\
      \textbf{KM-OR}& \textbf{.25} & \textbf{32.1 $\pm$ 0.6} & \textbf{39.7 $\pm$ 0.4} & \textbf{46.4 $\pm$ 0.2} \\
      AE & .75 & 46.8 $\pm$ 0.6 & 56.1 $\pm$ 0.3 & 58.9 $\pm$ 0.1 \\
      AE & .50 & 58.6 $\pm$ 0.3 & 60.0 $\pm$ 0.2 & 61.7 $\pm$ 0.1 \\
      \textbf{AE} & \textbf{.25} & \textbf{65.9 $\pm$ 0.6} & \textbf{71.6 $\pm$ 0.1} & \textbf{75.2 $\pm$ 0.4} \\
      \hline
      \multicolumn{2}{l}{\textbf{Supervised}}& & & \\
      CC-RS & .75 & 33.5 $\pm$ 0.2 & 42.0 $\pm$ 0.5 & 49.5 $\pm$ 0.4 \\
      \textbf{CC-RS} & \textbf{.50} & \textbf{35.2 $\pm$ 0.2} & \textbf{44.8 $\pm$ 0.3} & \textbf{51.9 $\pm$ 0.7} \\
      CC-RS* & .25 & 15.9 $\pm$ 0.7 & 17.7 $\pm$ 0.2 & 18.3 $\pm$ 0.1 \\
      \textbf{CC-OR} & \textbf{.75} & \textbf{66.6 $\pm$ 0.2} & \textbf{72.5 $\pm$ 0.2} & \textbf{75.7 $\pm$ 0.4} \\
      CC-OR & .50 & 57.0 $\pm$ 0.8 & 60.9 $\pm$ 0.7 & 64.2 $\pm$ 0.3 \\
      CC-OR & .25 & 58.6 $\pm$ 0.3 & 62.4 $\pm$ 0.1 & 66.6 $\pm$ 0.2 \\
      TL & .75 & 58.8 $\pm$ 0.3 & 60.0 $\pm$ 0.3 & 62.2 $\pm$ 0.5 \\
      \textbf{TL} & \textbf{.50} & \textbf{67.0 $\pm$ 0.2} & \textbf{73.1 $\pm$ 0.1} & \textbf{75.7 $\pm$ 0.2} \\
      TL & .25 & 49.0 $\pm$ 0.2 & 54.5 $\pm$ 0.4 & 54.7 $\pm$ 0.1 \\
      \hline
    \end{tabular}
\end{center}
\end{table}

\subsubsection{ENAS}
\autoref{table_enas} shows the Cell Evaluation for ENAS.
Unfortunately, ENAS did not perform well on all datasets, including the baseline. 
However, the evaluations on the proxy datasets reach similar results to the baseline in almost all experiments, which are also close to the results reported by Dong et al. \cite{dong2020bench}. The best cell design is surprisingly the design with only Skip-Connections. A cell design consisting of Skip-Connections and Average-Connections, where the averaging effect seems to lower the performance additionally, can explain the worse results. We observed that ENAS only chooses between those two operations, which will be examined in~\autoref{sec:op_dec} in more detail.

\begin{table}
\begin{center}
    \caption{Cell Evaluation with \textbf{ENAS}. The local optimum cell design (discussed in~\autoref{localOptimumCell}) is marked with *, which results in the best performing cell design. ENAS achieves only low accuracy compared to the other NAS approaches. Thus, there is no significant performance improvement or drop. All designs found on proxy datasets reach similar results compared to the baseline. }
    
    \label{table_enas}
    \begin{tabular}{l c c c c}
      \hline
      Method & $r$ & A, $C$=16 & A, $C$=32 & A, $C$=64 \\
      \hline
      \multicolumn{2}{l}{\textbf{Baseline}}& & & \\
      Full DS & 1.0 & 11.0 $\pm$ 0.5 & 12.5 $\pm$ 0.0 & 13.0 $\pm$ 0.0 \\
      \hline
      \multicolumn{2}{l}{\textbf{Unsupervised}}& & & \\
      RS & .75 & 11.7 $\pm$ 0.53 & 13.2 $\pm$ 0.1 & 13.4 $\pm$ 0.0 \\
      RS & .50 & 11.0 $\pm$ 0.50 & 12.5 $\pm$ 0.0 & 13.0 $\pm$ 0.0 \\
      \textbf{RS}& \textbf{.25} & \textbf{11.9 $\pm$ 0.35} & \textbf{13.8 $\pm$ 0.1} & \textbf{13.4 $\pm$ 0.0} \\
      KM-OR & .75 & 11.0 $\pm$ 0.5 & 12.5 $\pm$ 0.0 & 13.0 $\pm$ 0.0 \\
      \textbf{KM-OR} & \textbf{.50} & \textbf{11.5 $\pm$ 0.4} & \textbf{12.8 $\pm$ 0.1} & \textbf{13.1 $\pm$ 0.2} \\
      KM-OR & .25 & 11.4 $\pm$ 0.4 & 12.8 $\pm$ 0.1 & 13.1 $\pm$ 0.2 \\
      AE & .75 & 11.4 $\pm$ 0.4 & 12.8 $\pm$ 0.1 & 13.1 $\pm$ 0.2 \\
      AE & .50 & 11.0 $\pm$ 0.5 & 12.5 $\pm$ 0.0 & 13.0 $\pm$ 0.1 \\
      \textbf{AE} & \textbf{.25} & \textbf{11.4 $\pm$ 0.4} & \textbf{12.8 $\pm$ 0.1} & \textbf{13.1 $\pm$ 0.2} \\
      \hline
      \multicolumn{2}{l}{\textbf{Supervised}}& & & \\
      CC-RS & .75 & 10.6 $\pm$ 0.3 & 12.0 $\pm$ 0.1 & 12.6 $\pm$ 0.1 \\
      \textbf{CC-RS} & \textbf{.50} & \textbf{11.9 $\pm$ 0.4} & \textbf{13.2 $\pm$ 0.1} & \textbf{13.4 $\pm$ 0.0} \\
      CC-RS & .25 & 11.4 $\pm$ 0.4 & 12.8 $\pm$ 0.1 & 13.1 $\pm$ 0.2 \\
      %\hline
      CC-OR & .75 & 11.9 $\pm$ 0.4 & 13.2 $\pm$ 0.1 & 13.4 $\pm$ 0.0 \\
      \textbf{CC-OR*} & \textbf{.50} & \textbf{15.9 $\pm$ 0.7} & \textbf{17.7 $\pm$ 0.2} & \textbf{18.3 $\pm$ 0.1} \\
      CC-OR & .25 & 12.2 $\pm$ 0.4 & 13.3 $\pm$ 0.1 & 13.5 $\pm$ 0.1 \\
      \textbf{TL} & \textbf{.75} & \textbf{12.3 $\pm$ 0.4} & \textbf{13.4 $\pm$ 0.1} & \textbf{13.6 $\pm$ 0.1} \\
      TL & .50 & 11.7 $\pm$ 0.3 & 12.9 $\pm$ 0.1 & 13.2 $\pm$ 0.0 \\
      TL & .25 & 11.9 $\pm$ 0.4 & 13.2 $\pm$ 0.1 & 13.4 $\pm$ 0.0 \\
      \hline
    \end{tabular}
\end{center}
\end{table}

\subsubsection{GDAS}
\autoref{table_gdas} shows the Cell Evaluation for GDAS. 
It has a robust performance on proxies, which means that almost all experiments conclude similar performances to the baseline, except for $r=0.25$ (excluding CC-OR). Another exception is present for the proxy dataset KM-OR with $r=0.50$. Thus, GDAS benefits also from proxy datasets up to 50\%. Also, one can observe that the cell design with only Skip-Connections does not appear, which indicates that GDAS is more stable in search than DARTS or ENAS.

\begin{table}
\begin{center}
    \caption{Cell Evaluation with \textbf{GDAS}. There is no significant difference when comparing most experimental results with the baseline. Nevertheless, one can observe a performance drop for $r=0.25$. Moreover, the local optimum cell (discussed in~\autoref{localOptimumCell}) is not occuring for GDAS, which indicates a more stable NAS algorithm compared to DARTS and ENAS.}
    \label{table_gdas}
    \begin{tabular}{l c c c c}
    
      \hline
      Method & $r$ & A, $C$=16 & A, $C$=32 & A, $C$=64 \\
      \hline
      \multicolumn{2}{l}{\textbf{Baseline}}& & & \\
      Full DS & 1.0 & 65.8 $\pm$ 0.3 & 71.6 $\pm$ 0.3 & 74.3 $\pm$ 0.5 \\
      \hline
      \multicolumn{2}{l}{\textbf{Unsupervised}}& & & \\
      RS & .75 & 66.2 $\pm$ 0.0 & 71.3 $\pm$ 0.3 & 73.8 $\pm$ 0.2 \\
      \textbf{RS} & \textbf{.50} & \textbf{66.3 $\pm$ 0.3} & \textbf{71.1 $\pm$ 0.1} & \textbf{74.4 $\pm$ 0.3} \\
      RS & .25 & 47.7 $\pm$ 0.1 & 57.0 $\pm$ 0.4 & 60.4 $\pm$ 0.5 \\
      \textbf{KM-OR} & \textbf{.75} & \textbf{65.9 $\pm$ 0.8} & \textbf{70.9 $\pm$ 0.2} & \textbf{74.2 $\pm$ 0.4} \\
      KM-OR & .50 & 60.2  $\pm$ 0.3 & 65.6 $\pm$ 0.5 & 70.4 $\pm$ 0.2 \\
      KM-OR & .25 & 60.4 $\pm$ 0.6 & 66.0 $\pm$ 0.8 & 70.5 $\pm$ 0.4 \\
      \textbf{AE} & \textbf{.75} & \textbf{65.1 $\pm$ 0.6} & \textbf{69.8 $\pm$ 0.9} & \textbf{73.8 $\pm$ 0.4} \\
      AE & .50 & 64.2 $\pm$ 0.4 & 69.4 $\pm$ 0.2 & 73.0 $\pm$ 0.6 \\
      AE & .25 & 57.4 $\pm$ 0.8 & 63.9 $\pm$ 0.4 & 68.7 $\pm$ 0.5 \\
      \hline
      \multicolumn{2}{l}{\textbf{Supervised}}& & & \\
      \textbf{CC-RS} & \textbf{.75} & \textbf{65.8 $\pm$ 0.3} & \textbf{71.6 $\pm$ 0.3} & \textbf{74.3 $\pm$ 0.5} \\
      CC-RS & .50 & 65.7 $\pm$ 0.1 & 71.0 $\pm$ 0.3 & 74.1 $\pm$ 0.8 \\
      CC-RS & .25 & 30.6 $\pm$ 1.0 & 39.0 $\pm$ 0.8 & 45.9 $\pm$ 0.3 \\
      \textbf{CC-OR} & \textbf{.75} & \textbf{66.7 $\pm$ 0.5} & \textbf{71.5 $\pm$ 0.2} & \textbf{75.4 $\pm$ 0.2} \\
      CC-OR & .50 & 65.9 $\pm$ 0.2 & 71.4 $\pm$ 0.3 & 75.1 $\pm$ 0.5 \\
      CC-OR & .25 & 64.8 $\pm$ 0.2 & 70.7 $\pm$ 0.4 & 74.6 $\pm$ 0.2 \\
      TL & .75 & 66.0 $\pm$ 0.5 & 71.5 $\pm$ 0.4 & 74.2 $\pm$ 0.1 \\
      \textbf{TL} & \textbf{.50} & \textbf{66.4 $\pm$ 0.3} & \textbf{71.7 $\pm$ 0.4} & \textbf{75.0 $\pm$ 0.4} \\
      TL & .25 & 54.2 $\pm$ 0.0 & 62.4 $\pm$ 0.6 & 65.6 $\pm$ 0.1 \\
      \hline
    \end{tabular}
\end{center}
\end{table}

\subsubsection{Time Savings}
\begin{figure}
\begin{center}
    \includegraphics[width=1\linewidth]{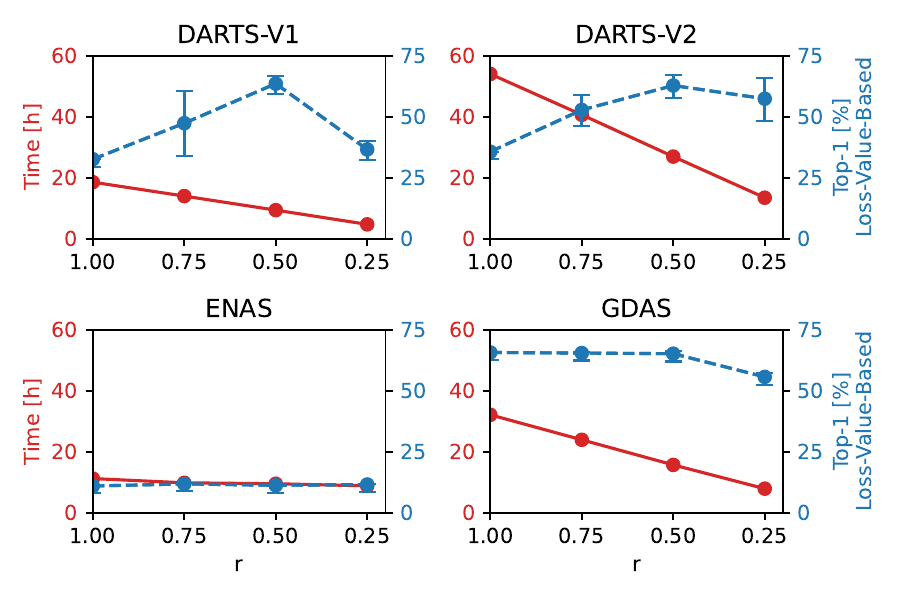}
\end{center}
\caption{Time savings and Accuracy results. The search time during Cell Search (all experiments) decreases with the dataset size. The standard deviation in time saving is not displayed because it is close to zero. In addition, top-1 accuracy mean and standard deviation for loss-value-based (AE \& TL) sampling is plotted. Interestingly, the accuracy improves for DARTS with decreasing $r$.}
\label{fig:timesavings}
\end{figure}

The search time during the experiments was measured with a system setup of a GPU model RTX 2080 by NVIDIA and a CPU model i9-9820X by Intel. \autoref{fig:timesavings} shows the time savings and Top-1 accuracy over $r$. 
We can observe a linear dependency between search time and sampling size.
Interestingly, ENAS does not profit like other approaches because the time savings are present for the controller training, not the child model's training, see Pham et al. \cite{pham2018efficient}. 
DARTS-V2 has the most significant gap between baseline and the time needed for the most reduced proxy dataset ($r=0.25$) with around 41 hours.
Thus, it is possible to derive a superior performing cell design with Cell Search and Cell Evaluation in one day with this setup.
In addition, it demonstrates that proxy datasets can improve the accuracy of the resulting architectures.
This is a very significant observation, especially for DARTS, where a long search time is necessary otherwise.

%-------------------------------------------------------------------------
\subsection{Qualitative Analysis}
This Section discusses the best performing cells, a local optimum cell design, and the operational decisions taken by NAS algorithms on the proxy datasets.

\subsubsection{Best Performing Cells}
\label{best_cells_sec}
\begin{figure}
\centering
\includegraphics[width=1\linewidth]{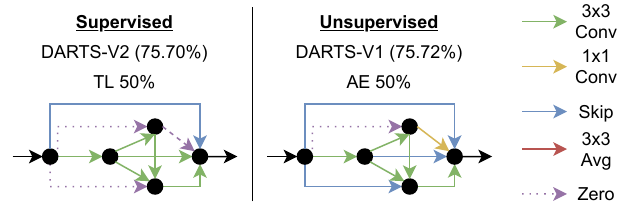}
\caption[Best performing cell designs]{Best performing cell designs for the unsupervised and supervised case. }
\label{fig:best_cells}
\end{figure}
Loss-value-based sampling methods (AE \& TL) found the best performing cell designs within proxy datasets, shown in~\autoref{fig:best_cells}. 
DARTS-V2 found it via the supervised sampling method TL ($r=0.50$) with 75.70\% accuracy and outperforms the baseline accuracy 53.74\% by +21.96 p.p. and GDAS (baseline) with 74.31\% accuracy, by +1.39 p.p., achieving a time saving of roughly 27.5 hours. 
Better by a small margin of +0.02 p.p. is the best cell design found via DARTS-V1 and the unsupervised sampling method AE ($r=0.50$), reaching 75.72\% top-1 accuracy. 
It comes with a time saving of ca. \textit{nine hours} and performance of +28,42 p.p. better than its baseline.

\begin{figure}
\centering
\includegraphics[width=.75\linewidth]{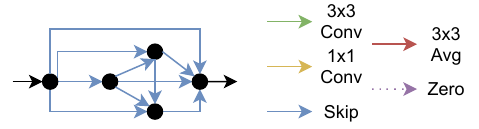}
\caption{Local optimum cell. One downside of proxy datasets is that many cell designs converge to Skip-Connections between the vertices for instable NAS algorithms like DARTS. Hence, the cell design does not contain any learnable parameter.}
\label{fig:c1}
\end{figure}

\subsubsection{Local Optimum Cell}
\label{localOptimumCell}
The experiments show that some Cell Searches obtained the same cell design, which yields abysmal accuracy. 
The most remarkable detail of this design is that the cell only uses Skip-Connections. 
Consequently, it does not contain any learnable parameter in the cell, explaining the lousy performance without further analysis. 
\autoref{fig:c1} illustrates the cell.
For DARTS, this is called the aggregation of Skip-Connections.
The local optimum cell problem is known \cite{chen2019progressive, liang2019darts+, bi2019stabilizing, zela2019understanding}. 
Unfortunately, there is no perfect solution to this problem up to this point. 
Nonetheless, proxy datasets seem to encourage this problem, which can be a benchmark for possible solutions.
This work's experiments show that proxy datasets can increase the search process's instability, especially for DARTS. 
On the other hand, the wide-reaching number of experiments show that this is not true for sample-based methods like GDAS. 
Thus, adding stochastic uncertainty seems to increase the stability.

\subsubsection{Generalizability}
In order to test the generalizability, we applied the best three cell designs (supervised and unsupervised, see~\autoref{best_cells_sec}) derived from CIFAR-100 to ImageNet-16-120, which is a down-sampled variant ($16 \times 16$) of 120 classes from ImageNet \cite{deng2009imagenet}.
\autoref{tab:imagenet16} lists the results.
Interestingly, a performance drop is observable by applying designs derived from CIFAR-100 to another dataset like ImageNet-16-120.
Nonetheless, the top-3 performing cell designs (supervised and unsupervised) still outperform the baseline except for one case (TL, $r=0.50$, GDAS).
However, we can conclude that searching on proxy datasets does not hurt the generalizability of found cell designs since the performance drop does not differ significantly from that of non-proxy datasets.
Moreover, like observed in~\autoref{localOptimumCell}, we can conclude that the bad experimental results of NAS-Bench-201 for DARTS are a product of the local optimum cell, which also explains the zero variance. This is because a cell design consisting only of Skip-Connections does not work better by applying different weight initializations.

\begin{table}
\begin{center}
    \caption{\label{tab:imagenet16}Top-1 accuracy of the top-3 best performing cells derived by loss-value-based sampling and applied on CIFAR-100 and ImageNet-16-120.  Note that the NAS approaches are trained on CIFAR-100 and are evaluated on ImageNet-16-120 similar to Dong et al. \cite{dong2020bench}. Also, our baseline uses a different macro skeleton than NAS-Bench-201 and  over-performance their results. The presented sampling methods (AE \& TL) reaches better results in both datasets than all baselines and ResNet using proxy datasets.}
    \begin{tabular}{lclcc}
    \hline
      & r & Model & CIFAR-100 & ImageNet \\
      & & & Acc [\%] & Acc [\%] \\

    \hline
    NAS-                     & 1.0 & \textbf{ResNet} & \textbf{70.9} & \textbf{43.6} \\ 
    Bench-                   & 1.0 & DARTS-V1 & 15.6 $\pm$ 0.0 & 16.3 $\pm$ 0.0 \\ 
    201 \cite{dong2020bench} & 1.0 & DARTS-V2 & 15.6 $\pm$ 0.0 & 16.3 $\pm$ 0.0 \\ 
                             & 1.0 & GDAS & 70.6 $\pm$ 0.3 & 41.8 $\pm$ 0.9 \\ 
    \hline
    Baseline & 1.0 & Darts-V1 & 47.3 $\pm$ 0.3 & $28.7 \pm 0.7$\\ %& 1.0 & GDAS & 74.3 $\pm$ 0.5 & $46.5 \pm 0.4$\\ 
    (ours)   & 1.0 & Darts-V2 & 53.7 $\pm$ 0.4 & $31.3 \pm 0.5$ \\ 
             & 1.0 & \textbf{GDAS} & \textbf{74.3} $\pm$ \textbf{0.5} & \textbf{46.5} $\pm \textbf{0.4}$\\ 
    \hline
    AE & .50 & \textbf{Darts-V1} & \textbf{75.7} $\pm$ \textbf{0.1} & \textbf{54.0} $\pm $ \textbf{0.5} \\  
     & .75 & GDAS & 73.8 $\pm$ 0.4 &  $47.9 \pm 1.0$\\
     & .50 & GDAS & 73.0 $\pm$ 0.6 & $48.1 \pm 0.2$\\
    \hline
    TL & .50 & \textbf{Darts-V2} & \textbf{75.7} $\pm$ \textbf{0.2} &  \textbf{54.2} $\pm $\textbf{0.8}\\
     & .50 & GDAS & 75.0 $\pm$ 0.4 &  $36.4 \pm 0.7$\\
     & .75 & GDAS & 74.2 $\pm$ 0.1 &  $48.0 \pm 0.5$\\
    \hline
    \end{tabular}
\end{center}
\end{table}

\subsubsection{Operation Decisions}
\label{sec:op_dec}
\begin{figure*}[th]
\centering
\includegraphics[height=.44\textheight,keepaspectratio]{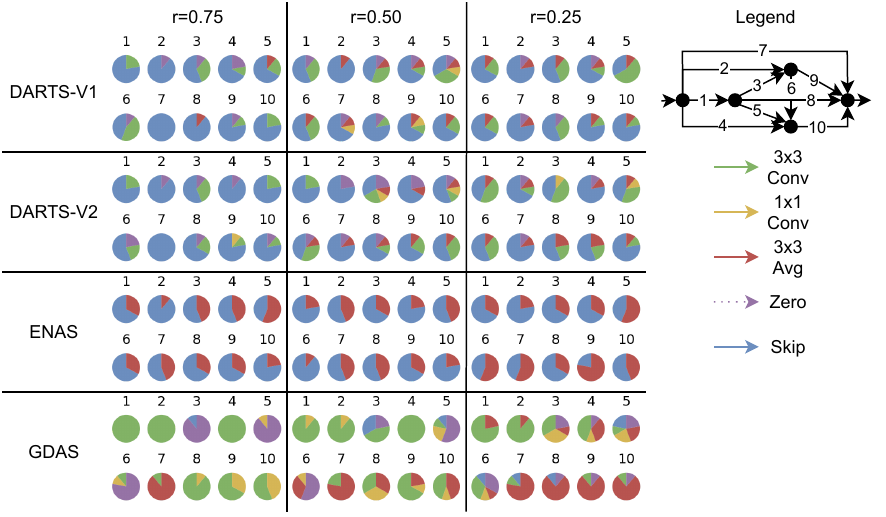}
\caption{Cell Edge Distribution. It contains the probability of an operation for a specific edge, following the notation on the bottom. The edges are numbered, and the operations are colored according to the legend on the right. Each cell design has ten edges, and therefore, there are ten pie charts for each entry. A strong dominance of Skip-Connections can be observed for DARTS and ENAS. In addition, ENAS only chooses between Skip- and Average-Connections, which explains the bad performance in the experiments. In contrast, GDAS uses more convolution operations. An exception is given for $r=0.25$, where Average-Operations become dominant. It aligns with the performance drop observed in the experiments.}
\label{fig:Celldistr}
\end{figure*}

This work was also interested in seeing how each NAS algorithm's operation choices change when applied to proxy datasets.
Thus, we derived an empirical probability distribution from all experiments and show the operations taken for each edge. 
This is done for all four algorithms to make a comparison feasible.
Consequently, it enables a distribution comparison with decreasing sampling size concerning operation choice. It will be referred to as Cell Edge Distribution in the following and is shown in~\autoref{fig:Celldistr}. To the best of our knowledge, this is the first work that uses this kind of visualization.

The distribution's most conspicuous findings are the colorings of ENAS.
It does not pick other operations than Averaging- or Skip-Connections (red and blue color). 
Thus, similar to the local optimum cell of~\autoref{localOptimumCell}, it does not contain any learnable parameter.
This argument is consistent $\forall r \in \{0.75, 0.50, 0.25\}$.
Consequently, it explains the low performance of ENAS throughout the experiments and its high robustness concerning the sampling methods tested.
Unfortunately, this raises the question of whether the NAS algorithm itself is bad on NAS-Bench-201 or if the implementation of the code framework of the authors of NAS-Bench-201 is not correct. 
For DARTS-V1 and V2, one can observe that both have similar decisions.
Also, a strong dominance of Skip-Connections (blue color) is present.
Concerning~\autoref{localOptimumCell}, the phenomenon of a cell design with only Skip-Connections extends to a general affinity to Skip-Connections.
GDAS has a significant difference from the other NAS approaches. 
It relies mainly on Convolution operations. Zeroize-Connections of edge 3, 5, and 6, which is very present for $r=0.75$ and partially for $r=0.50$ indicates a similarity for Inception-Modules (wide cell design) from GoogLeNet\cite{szegedy2015going}. 
As for DARTS, Average-Connections become more present for decreasing $r$, especially for the edges to the last vertex (7-10).
It is a possible explanation why GDAS underperforms for $r=0.25$ for almost all experiments.

\section{Conclusion \& Future Work}
%In this paper, \textit{Less is More} suggests that a well-defined reduced dataset enables a more efficient search procedure in terms of time and resulting design performance.
%While many other current methods exploit large datasets, this work advocates the opposite direction.
In this paper, we explored several sampling methods (supervised and unsupervised) for creating smaller proxy datasets, consequently reducing NAS approaches' search time.
Our evaluation is based on the prominent NAS-Bench-201 framework, adding the dataset ratio and different reduction techniques, resulting in roughly 1,400 experiments on CIFAR-100.
We further show the generalizability of the discovered architectures to ImageNet-16-120.
Within the evaluation, we find that many NAS approaches benefit from reduced dataset sizes (in contrast to current trends in research):
We find that not only the training time decreases linearly with the dataset reduction, but also that the accuracy of the resulting cells is oftentimes higher than when training on the full dataset.
Along those lines, DARTS-V2 found a cell design that achieves 75.7\% accuracy with only 25\% of the dataset, whereas the NAS baseline achieves only 53.7\% with all samples.
% Overall, depending on the algorithm, a time saving with at least similar results is achievable, e.g., DARTS-V2 in 13 instead of 54 hours. 
%Therefore, it is feasible with less data to derive more efficient designs that achieve higher performances.
%As overall, our findings show another direction regarding the role of the datasets in NAS approaches.
%For example, current methods in Self-Learning approaches exploit large unlabeled dataset, whereas this work suggest the opposite.
%In more detail, less is more refers that a high quality simplified dataset obtain many benefits in terms of time and performance.
%\textit{Less is More} suggests that a well-defined reduced dataset enables a more efficient search procedure in terms of time and resulting design performance.
% Generally, loss-value-based sampling works reliably well for the supervised and unsupervised case.
%Hence, the title \textit{Less is More} suggests that a well-defined reduced dataset enables a more efficient search procedure in terms of time and resulting design performance.
%While many other current methods exploit large datasets, this work advocates the opposite direction.
%However, we observe that NAS algorithms prone to instability (e.g., DARTS with the aggregation of Skip-Connections) could suffer even more on proxy datasets, e.g., Random Sampling.
Hence, overall reducing the size of the dataset is not only helpful to reduce the NAS search time but often also improves resulting accuracies (\textit{less is more}).

For future work, we observed that DARTS is more prone to instability in randomly sampled proxy datasets, which could also be presented for other NAS approaches and used as a benchmark to improve the stability.
Another direction for future work is to exploit synthetic datasets as proxy datasets, e.g., dataset distillation \cite{wang2018dataset}.

{\small
\bibliographystyle{ieee_fullname}
\bibliography{egbib}
}

\end{document}